\documentclass{article}

\pdfoutput=1

\PassOptionsToPackage{numbers, compress}{natbib}

\usepackage[dblblindworkshop, final]{neurips_2025}

\usepackage[pdftex]{graphicx}

\usepackage[utf8]{inputenc} 
\usepackage[T1]{fontenc}    
\usepackage{hyperref}       
\usepackage{url}            
\usepackage{booktabs}       
\usepackage{amsfonts}       
\usepackage{nicefrac}       
\usepackage{microtype}      
\usepackage{xcolor}         

\usepackage{comment}
\usepackage{pdfpages} 

\usepackage{url}

\usepackage{listings}
\usepackage{enumitem}
\setlist[enumerate]{left=0pt..1em} 
\setlist[itemize]{align=parleft,left=0pt..1em,topsep=0pt,parsep=0pt}   

\usepackage{subfig}  

\title{Reinforcement Learning for Long-Horizon Multi-Turn Search Agents}
\workshoptitle{NeurIPS 2025 Workshop on Multi-Turn Interactions in Large Language Models}

\author{%
  Vivek Kalyan \\
  Singapore \\
  \texttt{research@vivekkalyan.com} \\
  \And
  Martin Andrews \\
  Red Cat Labs, Singapore \\
  \texttt{martin@redcatlabs.com} \\
}

\begin{document}

\maketitle

\begin{abstract}
Large Language Model (LLM) agents can leverage multiple turns and tools to solve complex tasks, 
with prompt-based approaches achieving strong performance. 
This work demonstrates that Reinforcement Learning (RL) can push capabilities significantly further by learning from experience.
Through experiments on a legal document search benchmark, 
we show that our RL-trained 14 Billion parameter model outperforms frontier class models (85\% vs 78\% accuracy). 
In addition, we explore turn-restricted regimes, 
during training and at test-time, 
that show these agents achieve better results if allowed to operate over longer multi-turn horizons.
\end{abstract}

\section{Introduction}

Recent advances in LLM agents (\citet{Wang_2024,li-2025-review,li2025a})
have shown impressive capabilities in tool use (\citet{Qu2025-go}) 
and multi-step reasoning (\citet{wang-etal-2025-offline}). 
This has led to a growing interest in their application to complex, long-horizon interactive tasks
%
such as multi-turn document search, 
where an agent must interact with a document collection over several turns to locate specific information. 

Reinforcement Learning (RL, \citet{wen2025reinforcementlearningverifiablerewards}) 
offers a promising framework for training agents in these interactive settings.
The successful retrieval of a document provides a natural, verifiable reward signal 
that can be used to optimise the agent's behaviour programmatically. 
In this work, we explore the application of RL to multi-turn search agents in the legal domain.

Our key contributions are:
\begin{itemize}
\item Showing that on a legal dataset, a 14B RL-trained model is able to outperform frontier models accessible only through APIs; and
\item Exploring how RL-trained models can take advantage of the multi-turn setting 
      by running experiments in which the number of turns is restricted - both during training and at test-time.
\end{itemize}

\section{Related Work}

Retrieval Augmented Generation (RAG, \citet{10.5555/3495724.3496517, gao2024retrievalaugmentedgenerationlargelanguage}) 
has emerged as a promising solution for incorporating knowledge from external databases in the LLM era, 
helping to combat the challenges of hallucinations, outdated knowledge, and non-transparent, untraceable reasoning processes.
However, many implementations are either based around a single-retrieval phase \citet{lewis2020retrieval}, 
or a pre-set process to determine the retrievals (\citet{jiang2023active}).

Reinforcement Learning libraries, such as Agent Reinforcement Trainer (ART, \citet{hilton2025art}), 
simplify the RL training of tool use by LLM agents, 
allowing for techniques such as Chain of Retrieval (\citet{wang2025chainofretrievalaugmentedgeneration}) to be readily implemented.
In this work, we use ART to investigate the role of multi-turn behaviour for RAG tool use on a legal search benchmark.

\section{Methods}

\subsection{Task and dataset}

We construct a legal search benchmark from 5 years of Singapore court judgments. 
Each document in the dataset was parsed into an XML format, 
preserving its structural hierarchy with unique IDs for each section 
(e.g., \texttt{2021\_SGCA\_3:judgement:introduction:p1}).

Synthetic question-answer pairs were generated through a multi-stage pipeline: 
(i)   extracting patterns from seed queries provided by practicing lawyers;
(ii)  generating 10 candidate question/answer pairs per document using Gemini 2.5 Pro (\citet{comanici2025gemini25pushingfrontier}); and 
(iii) aggressively filtering these candidates based on criteria of realism, difficulty, and variety. 
The final dataset contains 2{\small,}300 questions with ground-truth documents, questions and answers.

\subsection{Na\"{\i}ve RAG baseline}

We established a single-turn RAG baseline where, given a query, the system: 
(i)   Executes searches using 
      (a) BM25 (\citet{Manning_Raghavan_Schütze_2008}) keyword search and
      (b) FAISS (\citet{douze2024faiss}) semantic search using the raw query; 
(ii)  Combines results from both methods; and  
(iii) Prompts the LLM to answer based solely on the retrieved context.

This baseline approach mirrors production RAG systems optimized for latency, 
making a single retrieval attempt without refinement or follow-up searches. 
Since a model cannot request additional information or explore related sections, 
this na\"{\i}ve RAG baseline forces it to work with whatever the initial retrieval returns.

\subsection{Models used in RL training}

The base model used for the RL training experiments was Qwen3-14B (\citet{qwen3}), 
whereas the 
Qwen3-0.6B model served as a low-cost alternative during code development, 
enabling the use of a much smaller GPU set-up.
To save on resources, only LoRA adapter (\citet{hu2022lora}) components were trained.
For the Reward Model, we used Gemini 2.5 Pro (\citet{comanici2025gemini25pushingfrontier}), 
which produced satisfactory binary decisions about the quality of the roll-out responses.

\subsection{Agent architecture}

All of the agents used in this work had access to three complementary tools for document exploration:
\begin{itemize}
\item \textbf{Keyword search}: Takes a query string and returns $K$ results from BM25 retrieval over document paragraphs. 
  Each result contains the section ID and a snippet highlighting the matched terms.

\item \textbf{Semantic search}: Takes a natural language query and returns $K$ results using cosine similarity 
over FAISS-indexed \texttt{all-MiniLM-L6-v2} (\citet{10.5555/3495724.3497138}) embeddings. 
  The returned results included section IDs and relevant snippets, enabling conceptual rather than lexical matching.

\item \textbf{Read document content}: Takes a section ID and returns the complete content of that section. 
  The hierarchical ID structure (e.g. \texttt{A:B:C}) also enables navigation: Agents can ``hop'' to parent sections by truncating IDs 
  (\texttt{A:B:C} $\rightarrow$ \texttt{A:B})
\end{itemize}

This system design creates a two-phase search pattern: broad exploration via keyword/semantic search to identify promising documents, 
followed by targeted reading to extract specific information.

The agent loop starts with the system prompt and query, 
and parses out \{\texttt{<think>}, \texttt{<tool>}, and \texttt{<answer>}\} sections from the response. 
Tool calls are executed and the results are returned to the model, continuing until the model produces an answer.

For details of the system prompt used for the Agentic settings, please see Appendix \ref{appendix-prompting}.

\subsection{Reinforcement Learning}

For the RL training, we used the ART library, which in turn used 
Parameter-Efficient Fine-Tuning (PEFT, \citet{peft}); 
\texttt{unsloth} from \citet{unsloth}; and the 
Transformer Reinforcement Learning library (\texttt{trl}, \citet{vonwerra2022trl})
to train model LoRA adapters. 

For each query, vLLM (\citet{kwon2023efficient}) was used to generate multiple trajectories from the model, 
from which RL rewards (detailed below) were calculated for each roll-out. 
YaRN (\citet{peng2024yarn}) was used to extend the context of vLLM to 128k tokens to allow for extended multi-turn roll-outs.
Following the reward evaluation, 
Group Relative Policy Optimization (GRPO, \citet{shao2024deepseekmathpushinglimitsmathematical}) 
was used to optimise the model policy. 
At the end of each step, the LoRA adapters used by vLLM were updated, 
so that the following roll-outs used the updated policy.
During training, \texttt{group\_size} was set to 6, and 8 groups were run per step.


A system of partial rewards 
was found necessary to enable Reinforcement Learning to work effectively.
Our reward structure created distinct behavioural bands that guided the model toward desired outcomes
by ensuring that even failed trajectories provided learning signal:

\begin{itemize}
\item \textbf{[1.0, 2.0]}: Correct answer with proper citations. Higher rewards for fewer turns/searches
\item \textbf{[0.0, 1.0]}: Model returns ``I don't know'' when unable to find sufficient evidence (preferable to hallucination)
\item \textbf{[-1.0, 0.0]}: Incorrect answer provided. Partial credit (+0.1 each) for finding correct documents
\item \textbf{[-2.0, -1.0]}: Formatting errors preventing tool execution (malformed tool calls, invalid arguments, non-existent document IDs)
\end{itemize}

Progress rewards (finding the right document, reading it, correct source citation) help the model learn intermediate skills necessary for the full task. 
Efficiency bonuses encourage models to achieve correct answers with fewer searches and turns. 
Critically, the above reward structure penalizes hallucination more severely than admitting uncertainty, 
training the model to say ``I don't know'' rather than fabricate answers when evidence is insufficient.

%

See Appendix \ref{appendix-metrics} for more information about the metrics used for tracking agents during RL-training.

\subsection{Turn-restricted evaluation}
\label{subsec:turn_restricted_eval}

To understand how models utilize multiple turns, 
we force early termination of the agent multi-turn tool-use
by prefixing \texttt{<answer>} to the assistant message at turn $N$, 
forcing the model to produce an answer.
%
Concretely, an $N$-turn roll-out has the following flow : 
\begin{center}
\texttt{query $\rightarrow$ response $\rightarrow$ \{reformulate search $\rightarrow$ response\}\textsuperscript{$\wedge$}$N$ $\rightarrow$ answer}
\end{center}
Thus, a $0$-turn rollout corresponds to na\"{\i}ve RAG.

\section{Results}

\subsection{Overall performance}

\begin{table}
  \caption{Performance comparison of multi-turn agents on legal document search}
  \label{tab:main-results}
  \centering
  \begin{tabular}{lcc}
    \toprule
    Model & Accuracy (\%) & Avg. Turns \\
    \midrule
    Na\"{\i}ve RAG (Gemini 2.5 Pro) & 33 & 1.0 \\
    Qwen3-14B (base) & 53 & 3.7 \\
    Gemini 2.5 Flash & 66 & 3.4 \\
    Gemini 2.5 Pro & 78 & 5.3 \\
    OpenAI o3 & 81 & 7.1 \\
    \textbf{Qwen3-14B + RL} & \textbf{85} & 6.2 \\
    \bottomrule
  \end{tabular}
\end{table}

Table \ref{tab:main-results} shows that the RL-trained Qwen3-14B achieves 85\% accuracy, 
surpassing all other models listed including frontier models that are only accessible via API. 
The progression from naive RAG (33\%) to multi-turn interaction shows clear benefits of iterative search, 
with each tier of model capability yielding substantial improvements.

The Qwen3-14B model without RL training plateaus at 53\% accuracy 
despite having access to the same tools and multi-turn interactions, 
highlighting that tool access alone is insufficient without learning how to use the capability effectively.

\begin{figure}
\centering
\begin{minipage}[t]{0.45\textwidth}
  \includegraphics[width=1.0\linewidth]{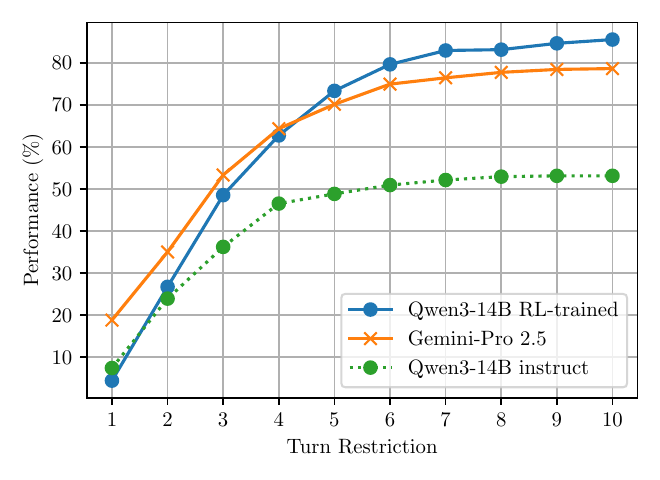}
  \caption{Performance of multi-turn agents under turn restrictions}
  \label{test-time-restrictions}
\end{minipage}%
\qquad
\begin{minipage}[t]{0.45\textwidth}
  \includegraphics[width=1.0\linewidth]{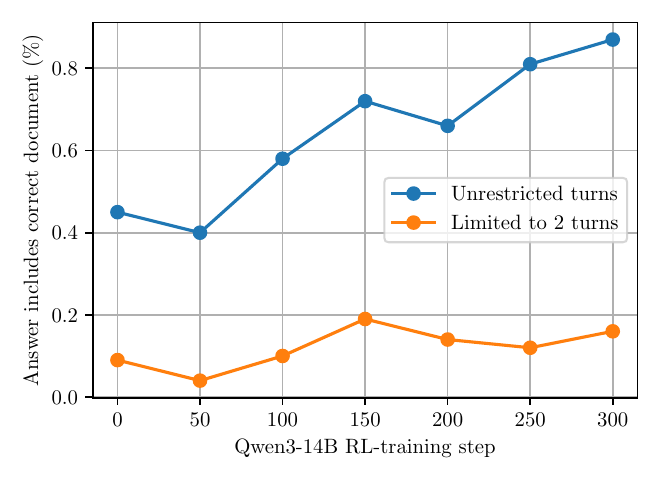}
  \caption{Effect of restricting turns during RL training}
  \label{train-time-restrictions}
\end{minipage}
\end{figure}

\subsection{Impact of turn-restricted inference}

To quantify the relationship between 
multi-turn interaction 
and model performance, 
we evaluated selected models under turn restrictions 
using the methodology described in Section \ref{subsec:turn_restricted_eval}.

Figure \ref{test-time-restrictions} shows the performances for each model under 1 to 10 turn restrictions.
All models exhibit monotonic improvement with additional turns, 
confirming that iterative search is essential for this task. 
The more significant finding is that 
the models diverge in their ability to exploit additional search opportunities :
The base Qwen3-14B performance plateaus after 6 turns, 
while the RL-trained variant and Gemini 2.5 Pro continue improving throughout all 10 turns.

In addition,
Gemini Pro 2.5 outperforms the RL-trained model in the low (less than 5) turn regime, 
whereas the far smaller RL-trained model shows additional benefit of multi-turn interactions beyond that. 
This phenomenon demonstrates a limitation of prompt-based approaches: 
while they can execute multi-turn search, 
they lack learned exploration strategies that compound value across multi-turn interactions.
This suggests that 
RL might play an important role in creating strong search agents.

\subsection{Impact of turn-restricted training}

To investigate whether effective multi-turn behaviour can be learned in turn-constrained settings, 
we also trained a Qwen3-14B model with a limit of 2 turns imposed during 
training, otherwise following the methodology of Section \ref{subsec:turn_restricted_eval}, 
with the plan of testing its generalisation capability later.

Figure \ref{train-time-restrictions} tracks the percentage of trajectories containing correct document citations 
across training steps for both the unrestricted and 2-turn restricted models. 
While the unrestricted model shows steady improvement from approximately 40\% to 85\% correct document identification over 300 training steps, 
the 2-turn restricted model exhibits no meaningful learning progress, 
fluctuating around its initial baseline of 10-15\% throughout training.

This failure to improve stems from the fundamental requirements of GRPO, 
which learns by comparing relative rewards within trajectory batches. 
With only 2 turns available, 
the model is unable to achieve correct answers at a high enough rate, 
providing insufficient positive examples for learning.

\section{Discussion}  




Reinforcement learning enables agents to learn effective tool use through practice rather than instruction. 
When agents encounter new tools or unfamiliar document collections, 
RL allows them to develop expertise through trial and error, 
discovering which search strategies work best. 
This approach is applicable to other tasks with multi-turn interactions, 
where agents must decide how to use their turns wisely, 
and this is an area of active, ongoing research.


\begin{ack}
%

This work's reinforcement learning experiments were significantly simplified by the use of the open-source ART library from OpenPipe. 
We acknowledge its crucial role in abstracting away low-level infrastructure tasks, including rollout management and weight updates.

Support for this research was provided by the Google AI Developer Programs team, 
including access to the Gemini models and GPUs on Google Cloud Platform.

The authors would also like to thank the reviewers for the NeurIPS 2025 Workshop on Multi-Turn Interactions in LLMs for their time and valuable feedback.

\end{ack}

\bibliographystyle{unsrtnat}  
\bibliography{main}

\begin{thebibliography}{24}
\providecommand{\natexlab}[1]{#1}
\providecommand{\url}[1]{\texttt{#1}}
\expandafter\ifx\csname urlstyle\endcsname\relax
  \providecommand{\doi}[1]{doi: #1}\else
  \providecommand{\doi}{doi: \begingroup \urlstyle{rm}\Url}\fi

\bibitem[Wang et~al.(2024)Wang, Ma, Feng, Zhang, Yang, Zhang, Chen, Tang, Chen, Lin, Zhao, Wei, and Wen]{Wang_2024}
Lei Wang, Chen Ma, Xueyang Feng, Zeyu Zhang, Hao Yang, Jingsen Zhang, Zhiyuan Chen, Jiakai Tang, Xu~Chen, Yankai Lin, Wayne~Xin Zhao, Zhewei Wei, and Jirong Wen.
\newblock A survey on large language model based autonomous agents.
\newblock \emph{Frontiers of Computer Science}, 18\penalty0 (6), March 2024.
\newblock ISSN 2095-2236.
\newblock URL \url{http://dx.doi.org/10.1007/s11704-024-40231-1}.

\bibitem[Li(2025{\natexlab{a}})]{li-2025-review}
Xinzhe Li.
\newblock A review of prominent paradigms for {LLM}-based agents: Tool use, planning (including {RAG}), and feedback learning.
\newblock In Owen Rambow, Leo Wanner, Marianna Apidianaki, Hend Al-Khalifa, Barbara~Di Eugenio, and Steven Schockaert, editors, \emph{Proceedings of the 31st International Conference on Computational Linguistics}, pages 9760--9779, Abu Dhabi, UAE, January 2025{\natexlab{a}}. Association for Computational Linguistics.
\newblock URL \url{https://aclanthology.org/2025.coling-main.652/}.

\bibitem[Li(2025{\natexlab{b}})]{li2025a}
Xinzhe Li.
\newblock A survey on {LLM} test-time compute via search: Tasks, {LLM} profiling, search algorithms, and relevant frameworks.
\newblock \emph{Transactions on Machine Learning Research}, 2025{\natexlab{b}}.
\newblock URL \url{https://openreview.net/forum?id=x9VQFjtOPS}.

\bibitem[Qu et~al.(2025)Qu, Dai, Wei, Cai, Wang, Yin, Xu, and Wen]{Qu2025-go}
Changle Qu, Sunhao Dai, Xiaochi Wei, Hengyi Cai, Shuaiqiang Wang, Dawei Yin, Jun Xu, and Ji-Rong Wen.
\newblock Tool learning with {LLMs}: {A} survey.
\newblock \emph{Front. Comput. Sci.}, 19\penalty0 (8), August 2025.

\bibitem[Wang et~al.(2025{\natexlab{a}})Wang, Hao, Dong, Zhang, Bao, Yang, and Wu]{wang-etal-2025-offline}
Huaijie Wang, Shibo Hao, Hanze Dong, Shenao Zhang, Yilin Bao, Ziran Yang, and Yi~Wu.
\newblock Offline reinforcement learning for {LLM} multi-step reasoning.
\newblock In Wanxiang Che, Joyce Nabende, Ekaterina Shutova, and Mohammad~Taher Pilehvar, editors, \emph{Findings of the Association for Computational Linguistics: ACL 2025}, pages 8881--8893, Vienna, Austria, July 2025{\natexlab{a}}. Association for Computational Linguistics.
\newblock ISBN 979-8-89176-256-5.
\newblock \doi{10.18653/v1/2025.findings-acl.464}.
\newblock URL \url{https://aclanthology.org/2025.findings-acl.464/}.

\bibitem[Wen et~al.(2025)Wen, Liu, Zheng, Xu, Ye, Wu, Liang, Wang, Li, Miao, Bian, and Yang]{wen2025reinforcementlearningverifiablerewards}
Xumeng Wen, Zihan Liu, Shun Zheng, Zhijian Xu, Shengyu Ye, Zhirong Wu, Xiao Liang, Yang Wang, Junjie Li, Ziming Miao, Jiang Bian, and Mao Yang.
\newblock Reinforcement learning with verifiable rewards implicitly incentivizes correct reasoning in base {LLMs}, 2025.
\newblock URL \url{https://arxiv.org/abs/2506.14245}.

\bibitem[Lewis et~al.(2020{\natexlab{a}})Lewis, Perez, Piktus, Petroni, Karpukhin, Goyal, K\"{u}ttler, Lewis, Yih, Rockt\"{a}schel, Riedel, and Kiela]{10.5555/3495724.3496517}
Patrick Lewis, Ethan Perez, Aleksandra Piktus, Fabio Petroni, Vladimir Karpukhin, Naman Goyal, Heinrich K\"{u}ttler, Mike Lewis, Wen-tau Yih, Tim Rockt\"{a}schel, Sebastian Riedel, and Douwe Kiela.
\newblock Retrieval-augmented generation for knowledge-intensive {NLP} tasks.
\newblock In \emph{Proceedings of the 34th International Conference on Neural Information Processing Systems}, NIPS '20, Red Hook, NY, USA, 2020{\natexlab{a}}. Curran Associates Inc.
\newblock ISBN 9781713829546.

\bibitem[Gao et~al.(2024)Gao, Xiong, Gao, Jia, Pan, Bi, Dai, Sun, Wang, and Wang]{gao2024retrievalaugmentedgenerationlargelanguage}
Yunfan Gao, Yun Xiong, Xinyu Gao, Kangxiang Jia, Jinliu Pan, Yuxi Bi, Yi~Dai, Jiawei Sun, Meng Wang, and Haofen Wang.
\newblock Retrieval-augmented generation for {Large Language Models}: A survey, 2024.
\newblock URL \url{https://arxiv.org/abs/2312.10997}.

\bibitem[Lewis et~al.(2020{\natexlab{b}})Lewis, Perez, Piktus, Petroni, Karpukhin, Goyal, K{\"u}ttler, Lewis, Yih, Rockt{\"a}schel, et~al.]{lewis2020retrieval}
Patrick Lewis, Ethan Perez, Aleksandra Piktus, Fabio Petroni, Vladimir Karpukhin, Naman Goyal, Heinrich K{\"u}ttler, Mike Lewis, Wen-tau Yih, Tim Rockt{\"a}schel, et~al.
\newblock Retrieval-augmented generation for knowledge-intensive {NLP} tasks.
\newblock \emph{Advances in neural information processing systems}, 33:\penalty0 9459--9474, 2020{\natexlab{b}}.

\bibitem[Jiang et~al.(2023)Jiang, Xu, Gao, Sun, Liu, Dwivedi-Yu, Yang, Callan, and Neubig]{jiang2023active}
Zhengbao Jiang, Frank~F Xu, Luyu Gao, Zhiqing Sun, Qian Liu, Jane Dwivedi-Yu, Yiming Yang, Jamie Callan, and Graham Neubig.
\newblock Active retrieval augmented generation.
\newblock In \emph{Proceedings of the 2023 Conference on Empirical Methods in Natural Language Processing}, pages 7969--7992, 2023.

\bibitem[Hilton et~al.(2025)Hilton, Corbitt, Corbitt, Gandhi, William, Kovalenskyi, and Jones]{hilton2025art}
Brad Hilton, Kyle Corbitt, David Corbitt, Saumya Gandhi, Angky William, Bohdan Kovalenskyi, and Andie Jones.
\newblock {ART}: Agent reinforcement trainer.
\newblock \url{https://github.com/openpipe/art}, 2025.

\bibitem[Wang et~al.(2025{\natexlab{b}})Wang, Chen, Yang, Huang, Dou, and Wei]{wang2025chainofretrievalaugmentedgeneration}
Liang Wang, Haonan Chen, Nan Yang, Xiaolong Huang, Zhicheng Dou, and Furu Wei.
\newblock Chain-of-retrieval augmented generation, 2025{\natexlab{b}}.

\bibitem[{Gemini Team}(2025)]{comanici2025gemini25pushingfrontier}
{Gemini Team}.
\newblock Gemini 2.5: Pushing the frontier with advanced reasoning, multimodality, long context, and next generation agentic capabilities, 2025.
\newblock URL \url{https://arxiv.org/abs/2507.06261}.

\bibitem[Manning et~al.(2008)Manning, Raghavan, and Schütze]{Manning_Raghavan_Schütze_2008}
Christopher~D. Manning, Prabhakar Raghavan, and Hinrich Schütze.
\newblock \emph{Introduction to Information Retrieval}.
\newblock Cambridge University Press, 2008.

\bibitem[Douze et~al.(2024)Douze, Guzhva, Deng, Johnson, Szilvasy, Mazaré, Lomeli, Hosseini, and Jégou]{douze2024faiss}
Matthijs Douze, Alexandr Guzhva, Chengqi Deng, Jeff Johnson, Gergely Szilvasy, Pierre-Emmanuel Mazaré, Maria Lomeli, Lucas Hosseini, and Hervé Jégou.
\newblock The {Faiss} library, 2024.

\bibitem[Yang et~al.(2025)Yang, Li, Yang, Zhang, Hui, Zheng, Yu, Gao, Huang, Lv, Zheng, Liu, Zhou, Huang, Hu, Ge, Wei, Lin, Tang, Yang, Tu, Zhang, Yang, Yang, Zhou, Zhou, Lin, Dang, Bao, Yang, Yu, Deng, Li, Xue, Li, Zhang, Wang, Zhu, Men, Gao, Liu, Luo, Li, Tang, Yin, Ren, Wang, Zhang, Ren, Fan, Su, Zhang, Zhang, Wan, Liu, Wang, Cui, Zhang, Zhou, and Qiu]{qwen3}
An~Yang, Anfeng Li, Baosong Yang, Beichen Zhang, Binyuan Hui, Bo~Zheng, Bowen Yu, Chang Gao, Chengen Huang, Chenxu Lv, Chujie Zheng, Dayiheng Liu, Fan Zhou, Fei Huang, Feng Hu, Hao Ge, Haoran Wei, Huan Lin, Jialong Tang, Jian Yang, Jianhong Tu, Jianwei Zhang, Jianxin Yang, Jiaxi Yang, Jing Zhou, Jingren Zhou, Junyang Lin, Kai Dang, Keqin Bao, Kexin Yang, Le~Yu, Lianghao Deng, Mei Li, Mingfeng Xue, Mingze Li, Pei Zhang, Peng Wang, Qin Zhu, Rui Men, Ruize Gao, Shixuan Liu, Shuang Luo, Tianhao Li, Tianyi Tang, Wenbiao Yin, Xingzhang Ren, Xinyu Wang, Xinyu Zhang, Xuancheng Ren, Yang Fan, Yang Su, Yichang Zhang, Yinger Zhang, Yu~Wan, Yuqiong Liu, Zekun Wang, Zeyu Cui, Zhenru Zhang, Zhipeng Zhou, and Zihan Qiu.
\newblock Qwen3 technical report.
\newblock \emph{arXiv preprint arXiv:2505.09388}, 2025.

\bibitem[Hu et~al.(2022)Hu, Shen, Wallis, Allen-Zhu, Li, Wang, Wang, and Chen]{hu2022lora}
Edward~J Hu, Yelong Shen, Phillip Wallis, Zeyuan Allen-Zhu, Yuanzhi Li, Shean Wang, Lu~Wang, and Weizhu Chen.
\newblock Lo{RA}: Low-rank adaptation of {Large Language Models}.
\newblock In \emph{International Conference on Learning Representations}, 2022.
\newblock URL \url{https://openreview.net/forum?id=nZeVKeeFYf9}.

\bibitem[Song et~al.(2020)Song, Tan, Qin, Lu, and Liu]{10.5555/3495724.3497138}
Kaitao Song, Xu~Tan, Tao Qin, Jianfeng Lu, and Tie-Yan Liu.
\newblock {MPNet}: masked and permuted pre-training for language understanding.
\newblock In \emph{Proceedings of the 34th International Conference on Neural Information Processing Systems}, NIPS '20, Red Hook, NY, USA, 2020. Curran Associates Inc.
\newblock ISBN 9781713829546.

\bibitem[Mangrulkar et~al.(2022)Mangrulkar, Gugger, Debut, Belkada, Paul, and Bossan]{peft}
Sourab Mangrulkar, Sylvain Gugger, Lysandre Debut, Younes Belkada, Sayak Paul, and Benjamin Bossan.
\newblock {PEFT}: State-of-the-art parameter-efficient fine-tuning methods.
\newblock \url{https://github.com/huggingface/peft}, 2022.

\bibitem[Daniel~Han and team(2023)]{unsloth}
Michael~Han Daniel~Han and Unsloth team.
\newblock Unsloth, 2023.
\newblock URL \url{http://github.com/unslothai/unsloth}.

\bibitem[von Werra et~al.(2020)von Werra, Belkada, Tunstall, Beeching, Thrush, Lambert, Huang, Rasul, and Gallouédec]{vonwerra2022trl}
Leandro von Werra, Younes Belkada, Lewis Tunstall, Edward Beeching, Tristan Thrush, Nathan Lambert, Shengyi Huang, Kashif Rasul, and Quentin Gallouédec.
\newblock {TRL}: Transformer reinforcement learning.
\newblock \url{https://github.com/huggingface/trl}, 2020.

\bibitem[Kwon et~al.(2023)Kwon, Li, Zhuang, Sheng, Zheng, Yu, Gonzalez, Zhang, and Stoica]{kwon2023efficient}
Woosuk Kwon, Zhuohan Li, Siyuan Zhuang, Ying Sheng, Lianmin Zheng, Cody~Hao Yu, Joseph~E. Gonzalez, Hao Zhang, and Ion Stoica.
\newblock Efficient memory management for large language model serving with {PagedAttention}.
\newblock In \emph{Proceedings of the ACM SIGOPS 29th Symposium on Operating Systems Principles}, 2023.

\bibitem[Peng et~al.(2024)Peng, Quesnelle, Fan, and Shippole]{peng2024yarn}
Bowen Peng, Jeffrey Quesnelle, Honglu Fan, and Enrico Shippole.
\newblock Ya{RN}: Efficient context window extension of {Large Language Models}.
\newblock In \emph{The Twelfth International Conference on Learning Representations}, 2024.
\newblock URL \url{https://openreview.net/forum?id=wHBfxhZu1u}.

\bibitem[Shao et~al.(2024)Shao, Wang, Zhu, Xu, Song, Bi, Zhang, Zhang, Li, Wu, and Guo]{shao2024deepseekmathpushinglimitsmathematical}
Zhihong Shao, Peiyi Wang, Qihao Zhu, Runxin Xu, Junxiao Song, Xiao Bi, Haowei Zhang, Mingchuan Zhang, Y.~K. Li, Y.~Wu, and Daya Guo.
\newblock {DeepSeekMath}: Pushing the limits of mathematical reasoning in open language models, 2024.
\newblock URL \url{https://arxiv.org/abs/2402.03300}.

\end{thebibliography}

\newpage

\appendix

\section{Appendix}

\subsection{Prompting}
\label{appendix-prompting}

The System Prompt for non-thinking models was as follows:
{\footnotesize
\begin{lstlisting}

You are a legal research assistant that can search legal documents to answer questions.

You have access to the following tools:

- search_keyword(query: str, num: int) -> str: Search using keyword/BM25 search for exact term matches.
- search_semantic(query: str, num: int) -> str: Search using semantic/vector search for conceptual similarity.
- read_document_part(part_id: str) -> str: Read a document part by ID. Part IDs use hierarchical format (e.g., A:B:C). To access parent parts, remove the last segment (e.g. A:B:C -> parent is A:B).

You may call one tool per turn, for up to {max_turns} turns, before giving your final answer.

In each turn, you should analyze what information you need and respond with EITHER a tool call OR your final answer.

For tool calls, use this format:
<think>
[your reasoning for what to search for and why]
</think>
<tool>
{"name": "tool_name", "args": {"query": "search query"}}
</tool>

When you have enough information, give your final answer in this format:

<think>
[your reasoning for the answer]
</think>
<answer>
[your comprehensive answer citing the evidence you found or "I don't know" if you didn't get enough information]

<sources>
<source>doc_id_1</source>
</sources>
</answer>

\end{lstlisting}
}

For the 'Thinking models', such as o-3 and Gemini Pro, we simply omit the instructions about the usage of \texttt{<think>} tags, 
and the thinking budgets were set to default values.

\newpage
\subsection{Metrics}
\label{appendix-metrics}

We tracked 13 distinct metrics across four categories to comprehensively assess agent performance:

\textbf{Final Outcome Metrics:}
\begin{itemize}
\item \texttt{answer\_correct}: Whether the answer matches ground truth (evaluated via LLM judge)
\item \texttt{sources\_correct}: Whether cited documents match ground truth documents (verifiable)
\item \texttt{returned\_i\_dont\_know}: Whether the model explicitly states uncertainty (verifiable)
\item \texttt{attempted\_answer}: Whether the model provided any answer (verifiable)
\end{itemize}

\textbf{Progress Tracking:}
\begin{itemize}
\item \texttt{ever\_found\_right\_doc}: Whether correct document appeared in any search results (verifiable)
\item \texttt{ever\_read\_right\_doc}: Whether the model used the read tool on the correct document (verifiable)
\end{itemize}

\textbf{Formatting Errors:}
\begin{itemize}
\item \texttt{cant\_parse\_tool\_call}: Malformed JSON or missing required tags (verifiable)
\item \texttt{bad\_tool\_call\_name}: Invalid tool name specified (verifiable)
\item \texttt{bad\_tool\_call\_args}: Incorrect arguments for valid tool (verifiable)
\item \texttt{bad\_sources\_id}: Referenced non-existent document IDs (verifiable)
\end{itemize}

\textbf{Efficiency Metrics:}
\begin{itemize}
\item \texttt{num\_turns}: Total number of tool-use turns taken (verifiable)
\item \texttt{num\_searches}: Count of keyword/semantic searches executed (verifiable)
\item \texttt{ran\_out\_of\_turns}: Whether the turn limit was reached (verifiable)
\end{itemize}

Notably, 12 of 13 metrics are verifiable without requiring an external judge, 
enabling fast and deterministic evaluation during training. 
Only \texttt{answer\_correct} requires LLM evaluation, 
for which we use Gemini 2.5 Pro to provide a True/False binary classification.

\end{document}